\documentclass{article}
\usepackage{spconf,amsmath,epsfig}
\usepackage{mathrsfs}
\usepackage{multirow}
\usepackage{xcolor}
\usepackage{comment}
\usepackage{arydshln}
\usepackage{url}
\newcommand{\todo}{{ \textbf{TODO}}\ }
\newcommand{\mycomment}[1]{}

\let\OLDthebibliography\thebibliography
\renewcommand\thebibliography[1]{
  \OLDthebibliography{#1}
  \setlength{\parskip}{0pt}
  \setlength{\itemsep}{0pt plus 0.3ex}
}
\title{Efficient Multi-Resolution Fusion \\ for Remote Sensing Data with Label Uncertainty}
\name{Hersh Vakharia and Xiaoxiao Du\thanks{This material is based upon work supported by the National Science
Foundation under Grant IIS-2153171-CRII: III: Explainable Multi-Source Data Integration with Uncertainty.}}
\address{University of Michigan}
\begin{document}
%
\maketitle
\begin{abstract}
Multi-modal sensor data fusion takes advantage of complementary or reinforcing information from each sensor and can boost overall performance in applications such as scene classification and target detection. This paper presents a new method for fusing multi-modal and multi-resolution remote sensor data without requiring  pixel-level training labels, which can be difficult to obtain. Previously, we developed a Multiple Instance Multi-Resolution Fusion (MIMRF) framework that addresses label uncertainty for fusion, but it can be slow to train due to the large search space for the fuzzy measures used to integrate sensor data sources. We propose a new method based on binary fuzzy measures, which reduces the search space and significantly improves the efficiency of the MIMRF framework. We present experimental results on synthetic data and a real-world remote sensing detection task and show that the proposed MIMRF-BFM algorithm can effectively and efficiently perform multi-resolution fusion given remote sensing data with uncertainty.
\end{abstract}
%

\section{INTRODUCTION}
\label{sec:intro}
Two fundamental challenges exist with remote sensor data integration. First, existing optical sensors operate on various spatial, spectral, or temporal resolutions. They may also produce heterogeneous data representations, such as hyperspectral imagery on a pixel grid or LiDAR (Light Detection and Ranging) point clouds with geometric measurements \cite{hong2021overview}. It is not always feasible to convert all data to the same resolution or map to the same grid for fusion without introducing misalignment errors or losing accuracy. Second, standard supervised learning methods generally require accurate labels for each data point, which can be difficult to obtain in high volume. Moreover, raw sensor data often contains missing and imprecise measurements due to occlusion, environmental factors, and sensor noise, making uncertainties unavoidable when fusing real-world remote sensing data.

To address these challenges, we previously proposed a Multiple Instance Multi-Resolution Fusion (MIMRF) method \cite{du2020multiresolution} for integrating multi-modal and imprecisely labeled remote sensing data. It formulated the label uncertainty problem 
using Multiple Instance Learning (MIL) \cite{dietterich1997solving} and relied on the Choquet integral \cite{choquet1954theory}, a powerful non-linear aggregation tool to perform multi-resolution fusion. However, the computational complexity of the original MIMRF algorithm grows exponentially with the number of sensor sources to be fused, which limits its efficiency.
In this work, we propose an efficient alternative to the original MIMRF algorithm by incorporating binary fuzzy measures (BFMs) with the MIL framework 
to improve the efficiency while maintaining the effectiveness
of multi-resolution fusion. 
Results are presented on real-world remote sensing data to demonstrate the efficiency and effectiveness of the proposed fusion algorithm.


The remainder of the paper is organized as follows. 
Section~\ref{sec:mimrf} describes the MIMRF framework. Section~\ref{sec:methodology} presents our proposed efficient MIMRF with binary fuzzy measures. Section~\ref{sec:results} presents experimental results. Section~\ref{sec:conclusion} discusses the proposed method and conclusions.



\section{The MIMRF Framework}
\label{sec:mimrf}
The MIMRF framework \cite{du2020multiresolution} takes heterogeneous sensor data as inputs (e.g., hyperspectral imagery and LiDAR point cloud) and learns a set of real-valued variables (``fuzzy measures'' \cite{keller2016fundamentals}) that reflects the interactions among input sensor sources. Assume there are $S$ sources to be fused, the size of the fuzzy measure is $2^S$ as each fuzzy measure element corresponds to a subset of the sensor combinations. A monotonic and normalized (non-binary) fuzzy measure can take any real value between 0 and 1. The higher the fuzzy measure value, the more 
weighting it is placing on the combination of sensor sources. 
The MIMRF algorithm uses the Choquet integral \cite{choquet1954theory} to perform fusion based on the learned fuzzy measures. The advantage of the MIMRF framework is that it can work with uncertain and imprecise labels, where training labels are associated with groups of data points (called ``bags'' or superpixels \cite{wang2017superpixel}) instead of each pixel, which greatly reduces the need to individually label every data point during training.

 

\section{Efficient MIMRF with BFM (Proposed)}
\label{sec:methodology}
\subsection{Binary Fuzzy Measures (BFMs)}
\label{sec:bfm}
\vspace{-2mm}
In contrast to MIMRF, our proposed efficient MIMRF-BFM algorithm utilizes binary fuzzy measures (BFMs) to reduce the search space and greatly improve the efficiency. A BFM is defined as a real valued function that maps $2^S \rightarrow \{0, 1\}$. It satisfies $\mathscr{G}(\emptyset) = 0$ (empty set rule); $\mathscr{G}(S) = 1$ (normalization);  and
 $\mathscr{G}(A) \leq \mathscr{G}(B)$ if $A \subseteq B$ and $A, B \subseteq S$ (monotonicity). Different from a standard real-valued (non-binary) normalized fuzzy measure used in the previous work, the BFM only take values of $0$ or $1$, instead of $[0,1]$. Thus, for $S$ sensor sources to be fused, BFM only needs to  search and optimize over $\{0, 1\}^{2^S}$ instead of $[0, 1]^{2^S}$ for the real-valued fuzzy measures, which leads to a simpler representation, a finite search space and more efficient computation. 

\vspace{-2mm}

\subsection{Objective Function and Model Learning}
\label{sec:bfm}
\vspace{-2mm}
As discussed in the Introduction, one challenge with fusing remote sensor data is to accommodate the difference in resolution and modality among sensor inputs. Assume we are fusing hyperspectral imagery (HSI) with LiDAR point cloud data. Each pixel in HSI may correspond to multiple data points in the LiDAR point cloud (see Figure 3 in \cite{du2020multiresolution} for an illustration). Additionally, we assume pixel-level labels are not accurate enough during training and we can only leverage bag-level labels with uncertainty (\textit{i.e.}, training labels are provided per superpixel but not on a pixel-level, which is very common in remote sensing data due to ground sample distance, sensor accuracy, etc.). Thus, we account for these two levels of fusion uncertainty by writing the objective function:
\vspace{-2mm}
\begin{equation}
\begin{aligned}
\min_\mathscr{G} \, J &=\sum_{a=1}^{B^-} \max_{\mathcal{S}_{ai}^-  \in \mathscr{B}_a^-} \left(  \boxed{\min_{\mathbf{x}_{k}^-  \in \mathcal{S}_{ai}^-}   C_\mathscr{G}(\mathbf{x}_{k}^- )} - 0 \right)^2 \\
&+ \sum_{b=1}^{B^+} \min_{\mathcal{S}_{bj}^+  \in \mathscr{B}_b^+} \left( \boxed{\max_{\mathbf{x}_{l}^+  \in \mathcal{S}_{bj}^+} C_\mathscr{G}(\mathbf{x}_{l}^+)} -1\right)^2,  
\end{aligned}
\label{eq:minmaxobj}
\end{equation}
%
where $B^+$ is the total number of positive bags containing the target object or material we wish to detect  (label $1$), $B^-$ is the total number of negative bags  containing only non-target background information (label $0$), $\mathbf{x}_{k}^-$ is the $i^{th}$ instance in the $a^{th}$ negative bag and $\mathbf{x}_{l}^+$ is the $j^{th}$ instance in the $b^{th}$ positive bag, $C_\mathscr{G}$ is the Choquet integral (CI) fusion output computed based on the binary fuzzy measure $\mathscr{G}$, $ \mathscr{B}_a^-$ is the $a^{th}$ negative bag, and $\mathscr{B}_b^+$ is the $b^{th}$ positive bag, and $ \mathcal{S}$ denotes the set of all possible matching combinations
of the multi-resolution/multi-modal sensor outputs. By minimizing this objective function, we seek the unknown binary fuzzy measure $\mathscr{G}$ given all the training data points ($\mathbf{x}$) and bag-level labels. The first term  encourages all instances in negative bags to produce a fusion result of “0” (non-target)
and the second term encourages at least one set of data instance
in positive bags to have label “1” (target), which satisfies the MIL assumption accounting for label uncertainty given multi-resolution data. 
The structure of our objective function is similar to the standard MIMRF method (Eq.(7) in \cite{du2020multiresolution}), but we introduce the novel use of BFMs to compress the search space of the fuzzy measure 
for enhanced efficiency. An evolutionary algorithm \cite{du2019multiple} was used to train and optimize the BFMs. The proposed MIMRF-BFM algorithm automatically learns the non-linear interactions and relationships (as represented by the BFM values) among the input sensor data sources to produce an optimized fusion result.


\section{Experimental Results}
\label{sec:results}
\subsection{MUUFL Gulfport Hyperspectral and LiDAR Fusion}
\begin{figure}[h]
\includegraphics[width=\columnwidth]{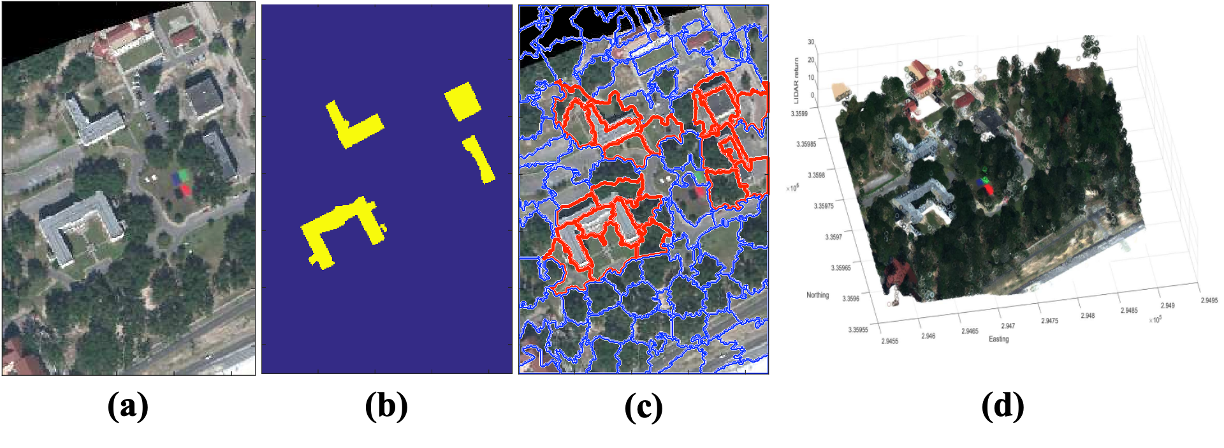}
\vspace{-10mm}
\caption{MUUFL Gulfport data. (a) Hyperspectral imagery. (b) Ground truth map for building detection (yellow represents the target buildings that we wish to detect); (c) Bag-level label map, where the superpixels containing target buildings are labeled positive (red) and the rest are labeled negative (blue). The superpixels are generated by the simple linear iterative clustering (SLIC) algorithm \cite{achanta2012slic}. (d) 3-D LiDAR point cloud.}
    \label{fig:muufl_gulfport_img}
\end{figure}

\vspace{-1mm}
Our proposed MIMRF-BFM algorithm was tested on a real-world remote sensing fusion task  on the MUULF Gulfport dataset \cite{alina_zare_2018_1186326}. The data set contains hyperspectral imagery and 3-D LiDAR point clouds collected during two aerial flights over the University of Southern Mississippi – Gulfpark campus. We follow a similar setup as in \cite{du2020multiresolution} and perform building detection by fusing three types of sensor inputs, one based on the adaptive coherence estimator (ACE) to detect building spectral signatures and two based on geometric measurements of building elevation from LiDAR point cloud data. These three sensor sources are multi-modal and multi-resolutional (multiple LiDAR points correspond to each HSI pixel due to the difference in sensor resolution and measurement inaccuracy). Figure~\ref{fig:muufl_gulfport_img} shows the MUUFL Gulfport data containing hyperspectral imagery, the LiDAR point clouds, and the ground truth map and superpixel-level labels for fusion task. 
\vspace{-2mm}
\vspace{-2mm}

\begin{figure}[ht!]
\centering
\includegraphics[width=0.94\columnwidth,trim={0 4mm 0 0},clip]{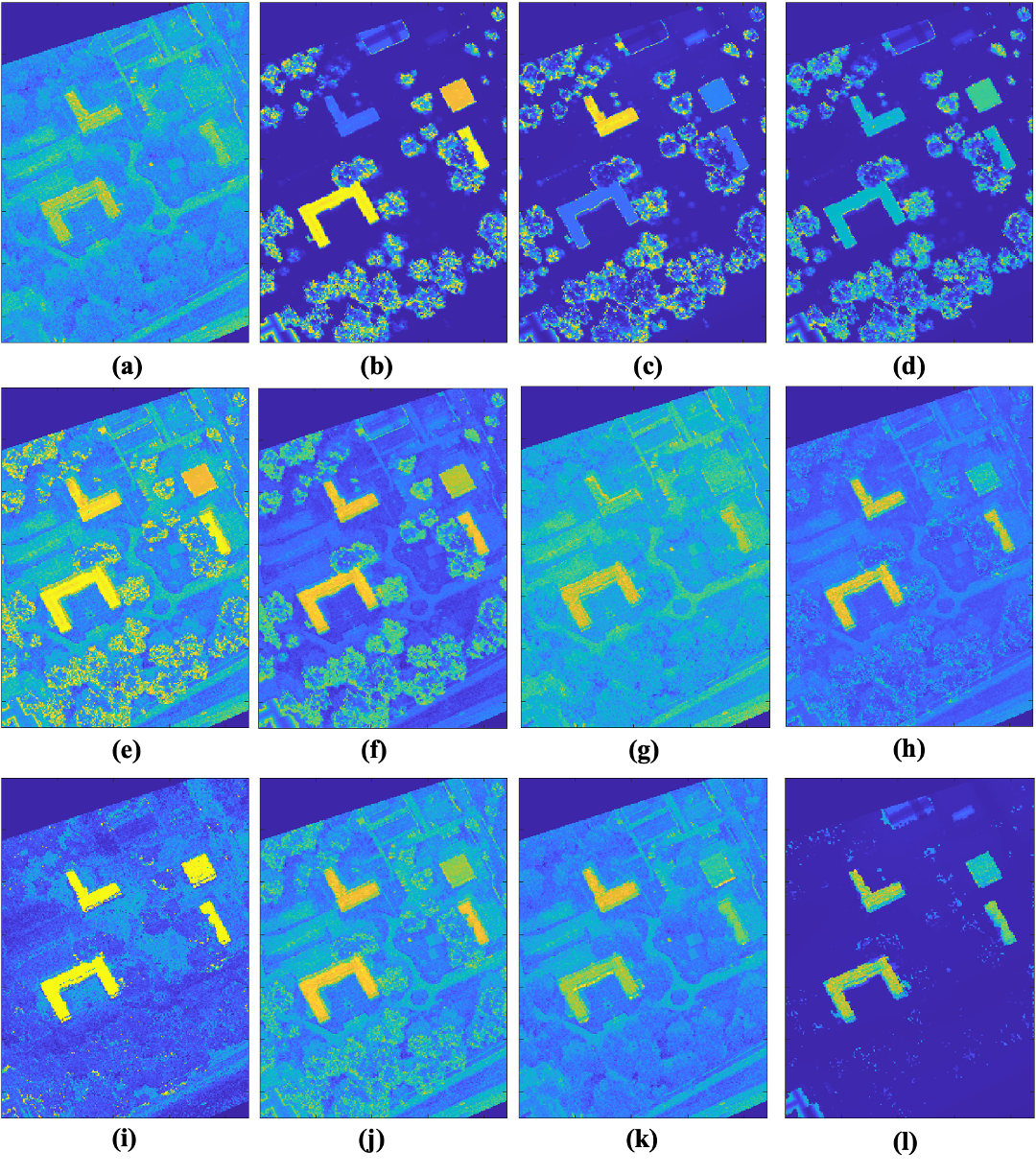}
\vspace{-3mm}
\caption{Confidence maps of the three individual sensor sources generated from the raw hyperspectral and LiDAR sensor data and visual results of various fusion techniques.  The color bar is between 0 and 1 (blue is low and yellow is high confidence). (a) ACE detector confidence based on the asphalt material in the hyperspectral imagery; (b)(c) Two LiDAR height maps that highlight different building structures; (d)(e)(f) Min, Max, and Mean of the three input sensor sources; (g) SVM; (h) mi-SVM; (i) KNN; (j) MICI \cite{du2018multiple} (non-multi-res); (k) MIMRF (multi-res, regular FM) \cite{du2020multiresolution}; and (l) Proposed MIMRF-BFM method (multi-res, with BFM).}
\label{fig:fuse_sqrt_all}
\end{figure}
\vspace{-5mm}

The proposed MIMRF-BFM algorithm was compared with the detection results from the three individual sensor input sources before fusion as well as a variety of other fusion methods to analyze its effectiveness and efficiency. The comparison methods include (i) non-multi-resolution fusion methods that requires pixel-perfect labels, such as taking the min/max/mean of the three input sources and using the support vector machine (SVM) and k-nearest neighbor (KNN) for classification; (ii) mi-SVM \cite{andrews2002support}, which is an MIL extension to SVM that works with bag-level labels; 
(iii) the Multiple Instance Choquet Integral (MICI) method with a noisy-or objective function \cite{du2016multiple,du2018multiple}, which is a non-multiresolution MIL fusion method based on the Choquet integral; (iv) previous MIMRF method with regular (non-binary) fuzzy measures; and (v) our proposed MIMRF-BFM method, which incorporates BFMs for added efficiency and works with both multi-resolution remote sensing data and bag-level training labels with uncertainty. 

\begin{table}[ht!]
\caption{The AUC and RMSE results of building detection using the MUUFL Gulfport data fusion (\textbf{Best}, \underline{Second Best}).}
\label{table:auc_ace_sqrt}
\centering
\resizebox{0.9\columnwidth}{!}{
\begin{tabular}{|l|c|c|c|c|}
\hline
\multicolumn{1}{|c|}{\multirow{2}{*}{\textbf{Fusion Method}}} & \multicolumn{2}{c|}{\textbf{AUC $\uparrow$ \, / \,  RMSE $\downarrow$ \, / \, PSNR $\uparrow$}} \\
\cline{2-3}
\multicolumn{1}{|c|}{} & \multicolumn{1}{l|}{\textbf{Train 1 Test 2}} & \multicolumn{1}
{l|}{\textbf{Train 2 Test 1}} \\
\hline
ACE& 0.906/0.362/8.839 & 0.952/0.346/9.214 \\
LiDAR1 & 0.888/0.267/11.497 & 0.880/0.272/11.319 \\
LiDAR2  & 0.850/0.273/11.243 & 0.839/0.280/11.053 \\
\hdashline
Min & 0.877/0.255/12.262 & 0.867/0.261/11.673 \\
Max & 0.916/0.434/7.333 & 0.932/0.422/7.501 \\
Mean & 0.941/0.310/10.492 & 0.953/0.302/10.400 \\
SVM & 0.892/0.415/7.637 & 0.958/0.285/7.637 \\
mi-SVM & 0.951/\underline{0.226}/12.379 & 0.972/\underline{0.203}/\underline{13.863} \\
KNN & 0.954/0.237/\underline{12.437} & 0.952/0.243/12.279 \\
MICI Noisyor & 0.943/0.377/8.621 & 0.946/0.326/9.030 \\
MIMRF & \textbf{0.976}/0.310/10.314 & \textbf{0.989}/0.254/10.635 \\
\textbf{MIMRF-BFM} & \underline{0.974}/\textbf{0.131}/\textbf{17.661} & \underline{0.973}/\textbf{0.128}/\textbf{17.859} \\
\hline
\end{tabular}}
\end{table}

\begin{figure}[ht!]
\includegraphics[width=\columnwidth,trim={0 0 0 0},clip]{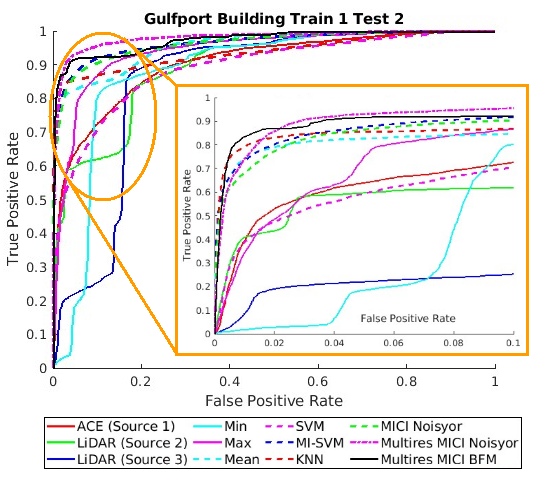}

\caption{ROC curve results of building detection accuracy on the MUUFL Gulfport dataset (cross validated over two flights). Orange box: zoomed-in view. Best viewed in color.}
    \label{fig:roc}
\end{figure}

Figure~\ref{fig:roc} shows a ROC (receiver operating characteristic) curve result and Table~\ref{table:auc_ace_sqrt} shows the AUC (area under curve), RMSE (root mean square error from ground truth), and PSNR (peak signal-to-noise ratio from ground truth) results across all methods for quantitative comparison. Higher AUC and PSNR and lower RMSE indicate better detection results after fusion. As shown, our proposed MIMRF-BFM achieved high detection performance, low error, and low noise overall.

MIMRF-BFM produced an AUC score second to non-binary MIMRF, but outperformed other fusion techniques. MIMRF-BFM  also produced the lowest RMSE and highest PSNR, indicating that the BFMs allow background noise to be eliminated more effectively.

Figure~\ref{fig:fuse_sqrt_all} shows visual results of the sensor inputs and the fusion maps. Compared to the non-binary MIMRF results (subfigure k), our proposed MIMRF-BFM (subfigure l) suppresses the background pixels much better and only assigns a high confidence score to the desired building pixels. This demonstrated that the BFM actually helps improve the contrast between positive and negative classes and benefits detection, as the learned BFM only takes values in $\{0,1\}$.

 We also investigated the actual BFM values learned during the fusion process. 
 The final BFM learned by the proposed MIMRF-BFM algorithm is $\mathscr{G}_{12}= \mathscr{G}_{13}=1$, and other measure elements ($\mathscr{G}_{1}$, $\mathscr{G}_{2}$, $\mathscr{G}_{3}$, $\mathscr{G}_{23}$) were zero. This means that the MIMRF-BFM algorithm correctly identified that the intersection between source 1\&2 and source 1\&3 contributed most to the detection of all four buildings. This makes sense, as the ACE detector (source 1) highlighted all asphalt materials including building rooftops as well as roads, whereas source 2 and 3 are LiDAR height maps which highlighted partial buildings but also tree canopies. By learning the combination among these sources, the MIMRF-BFM was able to suppress false detections of other materials such as roads and trees and instead only place high confidence on the targets (buildings). In terms of computation time, the MIMRF takes on average $30$min to train, whereas the MIMRF can complete the search and finish training in $10$s (over 180 times faster).


\subsection{Multi-Source BFM Efficiency Analysis}
To further analyze the efficiency of the proposed MIMRF-BFM approach, we generated an additional synthetic multi-resolution dataset with incrementally growing (6, 8, 10, and 12) number of sensor sources and compared the computation efficiency of the proposed MIMRF-BFM to the original MIMRF (non-BFM) algorithm. The computation time (the time it took to converge on an optimal FM/BFM) was recorded and presented in Table \ref{table:manysource}. It is clear that the proposed MIMRF-BFM algorithm is capable of learning how to fuse a high number of sources significantly faster than the original (non-binary) MIMRF. When the input sensor sources to be fused increased to over ten, the MIMRF without BFMs consistently require more than five hours to train without a guarantee to converge, whereas the MIMRF-BFM can finish significantly faster due to the reduction in search space.

\begin{table}[h!]
\caption{Computation time comparison between MIMRF and MIMRF-BFM. Mean(Standard deviation) across 5 runs. Computation was capped at 5 hours.}
\label{table:manysource}
\centering
\resizebox{\columnwidth}{!}{
\begin{tabular}{|l|c|c|c|c|}
\hline
\multicolumn{1}{|c|}{\multirow{2}{*}{\textbf{Fusion Method}}} & \multicolumn{4}{|c|}{\textbf{Computation Time (s)}} \\
\cline{2-5}
\multicolumn{1}{|c|}{} & \multicolumn{1}{c|}{\textbf{\#6}} & \multicolumn{1}{c|}{\textbf{\#8}} & \multicolumn{1}{c|}{\textbf{\#10}} & \multicolumn{1}{c|}{\textbf{\#12}}\\
\hline
MIMRF & 149.5(148.0) & 772.1(442.1) & \textgreater{}$5$ hrs. & \textgreater{}$5$ hrs.\\
\textbf{MIMRF-BFM} & 17.6(1.1) & 92.1(5.1) & 120.3(5.0) & 772.4(15.9)\\
\hline
\end{tabular}}
\end{table}

\vspace{-3mm}

\section{Discussion and Conclusions}
\label{sec:conclusion}
This paper presents MIMRF-BFM, an effective and efficient extension to the previously developed MIMRF framework that incorporates binary fuzzy measures. 

The use of BFMs drastically reduced the search space during model learning and  resulted in significant improvements in efficiency, which was particularly useful when scaling up the number of fusion sources.  Additionally, the learned BFM measure values provide a clear and explainable representation corresponding to the combination and (non-linear) interactions of the sensor input sources, which allows humans to interpret and gain insights on the fusion process.

In addition to the HSI and LiDAR fusion task presented in this paper, the proposed MIMRF-BFM can be extended to fusing alternative sensor modalities accounting for varying resolution and label uncertainties. Future work includes adapting the MIMRF-BFM to multi-view, multi-temporal and multi-spatial sensor data with uncertainty and incorporating generalized fuzzy measures, such as fuzzy measures on a bipolar scale (e.g., when FMs map to $[-1,1]$) \cite{grabisch2005bi}.

\vspace{-1mm}

\bibliographystyle{IEEEbib}
\small
\bibliography{references.bib}

\end{document}